\documentclass[lettersize,journal]{IEEEtran}
\usepackage{amsmath,amsfonts}
\usepackage{algorithmic}
\usepackage{algorithm}
\usepackage{array}
\usepackage[caption=false,font=normalsize,labelfont=sf,textfont=sf]{subfig}
\usepackage{textcomp}
\usepackage{stfloats}
\usepackage{url}
\usepackage{verbatim}
\usepackage{graphicx}
\usepackage{cite}

\usepackage{times}
\usepackage{epsfig}
\usepackage{amsmath}
\usepackage{amssymb}
\usepackage{eso-pic}

\usepackage{lineno}
\usepackage{booktabs,makecell, multirow, tabularx}
\usepackage{tikz,xcolor}
\usepackage{hyperref}

\hypersetup{hidelinks,
	colorlinks=true,
	allcolors=black,
	pdfstartview=Fit,
	breaklinks=true}

\begin{document}	
\title{Domain Consistency Representation Learning for Lifelong Person Re-Identification}
\author{
Shiben Liu \href{https://orcid.org/0000-0001-9376-2562}{\includegraphics[scale=0.08]{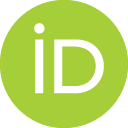}}, 
Huijie Fan
\href{https://orcid.org/0000-0002-8548-861X}{\includegraphics[scale=0.08]{ORCIDiD_icon128x128.png}},
Qiang Wang \href{https://orcid.org/0000-0002-2018-1764}{\includegraphics[scale=0.08]{ORCIDiD_icon128x128.png}}, 
Weihong Ren \href{https://orcid.org/0000-0003-4756-3962}{\includegraphics[scale=0.08]{ORCIDiD_icon128x128.png}}, 
Yandong Tang \href{https://orcid.org/0000-0003-3805-7654}{\includegraphics[scale=0.08]{ORCIDiD_icon128x128.png}},
Yang Cong
\href{https://orcid.org/0000-0002-5102-0189}{\includegraphics[scale=0.08]{ORCIDiD_icon128x128.png}, ~\IEEEmembership{Senior Member, IEEE}}
\thanks{This work is supported by the National Natural Science Foundation of China (62273339, U24A201397),  the Key Research and Development Program of Liaoning (2024JH2/102400022) and the LiaoNing Revitalization Talents Program (XLYC2403128). (\emph{Corresponding author: Huijie Fan})}
\thanks{Shiben Liu is with the State Key Laboratory of Robotics, Shenyang Institute of Automation, Chinese Academy of Sciences, Shenyang 110016, China, and also with the University of Chinese Academy of Sciences, Beijing 100049, China (e-mail: liushiben@sia.cn).
\par
Huijie Fan, and Yandong Tang are with the State Key Laboratory of Robotics, Shenyang Institute of Automation, Chinese Academy of Sciences, Shenyang, 110016, China (e-mail: fanhuijie@sia.cn; ytang@sia.cn).
\par 
Qiang Wang is with the Key Laboratory of Manufacturing Industrial Integrated Automation, Shenyang University, and with the State Key Laboratory of Robotics, Shenyang Institute of Automation, Chinese Academy of Sciences, Shenyang, 110016, China (e-mail: wangqiang@sia.cn). 
\par 
Weihong Ren is with the Harbin Institute of Technology, Shenzhen 518055, China (e-mail: renweihong@hit.edu.cn).
\par
Yang Cong is with the College of Automation Science and Engineering, South China University of Technology, Guangzhou, 510640, China (e-mail: congyang81@gmail.com).
\par
}
}


\maketitle
\begin{abstract}
Lifelong person re-identification (LReID) exhibits a contradictory relationship between intra-domain discrimination and inter-domain gaps when learning from continuous data. Intra-domain discrimination focuses on individual nuances ($\textit{i.e.}$, clothing type, accessories, $\textit{etc.}$), while inter-domain gaps emphasize domain consistency. Achieving a trade-off between maximizing intra-domain discrimination and minimizing inter-domain gaps is a crucial challenge for improving LReID performance. Most existing methods strive to reduce inter-domain gaps through knowledge distillation to maintain domain consistency. However, they often ignore intra-domain discrimination. To address this challenge, we propose a novel domain consistency representation learning (DCR) model that explores global and attribute-wise representations as a bridge to balance intra-domain discrimination and inter-domain gaps. At the intra-domain level, we explore the complementary relationship between global and attribute-wise representations to improve discrimination among similar identities. Excessive learning intra-domain discrimination can lead to catastrophic forgetting. We further develop an attribute-oriented anti-forgetting (AF) strategy that explores attribute-wise representations to enhance inter-domain consistency, and propose a knowledge consolidation (KC) strategy to facilitate knowledge transfer. Extensive experiments show that our DCR achieves superior performance compared to state-of-the-art LReID methods. Our code is available at \url{https://github.com/LiuShiBen/DCR}. 
\end{abstract}

\begin{IEEEkeywords}
Lifelong person re-identification, attribute-text generator, text-image aggregation, domain consistency representation.
\end{IEEEkeywords}

\section{Introduction}
\IEEEPARstart{P}{erson} re-identification (ReID) aims to retrieve the same individual across multiple cameras in a large-scale database by using uni-modal architectures such as convolutional neural networks (CNN) \cite{zhang2021coarse, pang2023cross, ran2024multi} or vision transformers (ViT) \cite{chen2023hasi, yang2024pedestrian, fan2023skip}. However, when ReID models are applied to continuous datasets collected by video-based monitoring systems \cite{tian2024novel, liu2024wavelet}, they exhibit notable performance limitations. As a result, recent works have focused on the practical problem of lifelong person identification (LReID), which maintains strong performance with continuously updated data streams.\\
\begin{figure}[t]
	\centering 
	\includegraphics[width=1.0\linewidth, height=0.43\textheight]{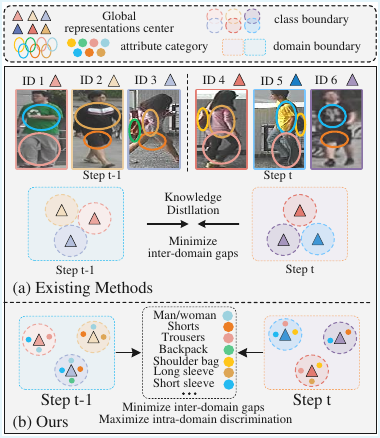}
	\caption{Comparison between our method and existing methods. (a) Existing methods \cite{ge2022lifelong,yu2023lifelong} leverage knowledge distillation to minimize inter-domain gaps but ignore intra-domain discrimination, which limits the LReID model's ability to learn new knowledge. (b) Our method explores domain consistency representations as a bridge to achieve a trade-off between maximizing intra-domain discrimination and minimizing inter-domain gaps, enhancing the LReID model's anti-forgetting and generalization capabilities.}
	\label{fig:fig1}
\end{figure}
\indent At present, lifelong person re-identification (LReID) suffers from the challenge of balancing the anti-forgetting of old knowledge and learning new knowledge. Specifically, there are two main issues to solve this challenge. 1) \textbf{Intra-domain discrimination}. Each identity may exhibit subtle nuances of individual information ($\textit{i.e.}$, clothing type, accessories, haircut, $\textit{etc.}$) and lead to severe distribution overlapping. Learning discriminative representations of individuals are effective for distinguish identity information. 2) \textbf{Inter-domain gaps}. Each Domain is collected in different illumination and background, leading to inter-domain gaps. Bridging intra-domain gaps are significant for mitigating catastrophic forgetting in LReID.  \\
\indent To address these issues, we aim to learn domain consistency representations that capture individual nuances in intra-domain and inter-domain consistency in LReID. Knowledge distillation-based approaches \cite{yu2023lifelong, pu2022meta, huang2023learning} ensure distribution consistency between the previous and current domain to alleviate catastrophic forgetting. However, these approaches impose strict constraints and ignore intra-domain discrimination, \cite{wu2021generalising, zhang2023spatial, xu2024lstkc}, as outlined in Fig. \ref{fig:fig1}(a). While LReID models significantly improve intra-domain discrimination for the current step, they inevitably damage inter-domain consistency, leading to catastrophic forgetting. Thus, we explore global and attribute-wise representations to strike a trade-off between maximizing intra-domain discrimination and minimizing inter-domain gaps, improving the anti-forgetting and generalization capabilities of the LReID model, as illustrated in Fig. \ref{fig:fig1}(b). \\
\indent Specifically, we propose a novel domain consistency representation learning (DCR) model that first explores attribute and text information to enhance LReID performance. Unlike methods \cite{sun2022patch, xu2024distribution, cui2024learning}, we develop domain consistency representations including global and attribute-wise representations to capture individual nuances in intra-domain and inter-domain consistency in LReID. We design an attribute-text generator (ATG) to dynamically generate text-image pairs for each instance, which are then fed into a text-guided aggregation (TGA) network to enhance the global representation capability, effectively distinguishing identities in LReID. Furthermore, the attributes of each instance guide an attribute compensation (ACN) network to generate attribute-wise representations that focus on specific regional information about identities. We consider that attributes can enhance reliability by setting higher thresholds across domains. Therefore, the generated attribute-wise representations and text for each instance are considered reliable in our model.\\
\indent In summary, we aim to strike a balance between maximizing intra-domain identity-discriminative information and minimizing inter-domain gaps by exploring global and attribute-wise representations. At the intra-domain level, global representations capture whole-body information, while attribute-wise representations focus on specific regional information. When whole-body appearances or attribute-related information are similar across identities, we combine global and attribute-wise representations to distinguish among similar identities, maximizing intra-domain discrimination. While the model adapts well to new information, it often struggles with catastrophic forgetting. To preserve old knowledge, we develop an attribute-oriented anti-forgetting (AF) strategy that explores attribute-wise representations to bridge inter-domain gaps across continuous datasets. Meanwhile, striking a trade-off between maximizing intra-domain discrimination and minimizing inter-domain gaps is crucial for improving the performance of LReID models. Knowledge consolidation (KC) is proposed to facilitate knowledge transfer and enhance generalization capabilities, which consists of alignment and logit-level distillation mechanisms. The alignment mechanism explores global representations of knowledge transfer from the current dataset. The logit-level distillation mechanism enhances the extraction of identity information, thereby improving the model's ability to consolidate knowledge. Our contributions are as follows:\\

\begin{itemize}
	\item We propose a novel domain consistency representation learning (DCR) model that explores global and attribute-wise representations to capture individual nuances in intra-domain and inter-domain consistency, achieving a trade-off between maximizing intra-domain discrimination and minimizing inter-domain gaps.
	\item In the intra-domain context, we explore the complementary relationship between global and attribute-wise representations to enhance the discrimination of each identity and adapt to new knowledge.
	\item In the inter-domain context, we design an attribute-oriented anti-forgetting (AF) and a knowledge consolidation (KC) strategy to minimize inter-domain gaps and facilitate knowledge transfer, improving the LReID model’s generalization and anti-forgetting capabilities.
\end{itemize} 
\section{Related work}
\begin{figure*}[!t]
	\centering 
	\includegraphics[width=1.0\linewidth, height=0.49\textheight]{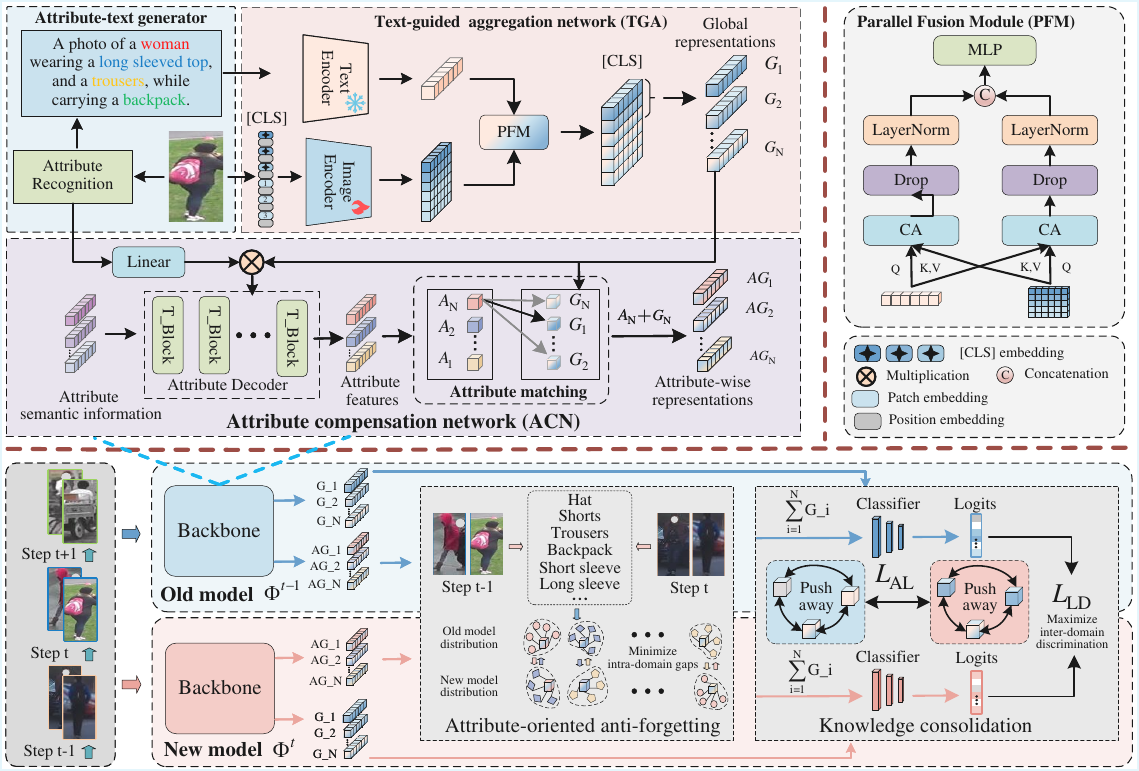}
	\caption{Overview of the proposed DCR for LReID. First, the attribute-text generator (ATG) dynamically generates text-image pairs for each instance. Then, the text-guided aggregation network (TGA) captures global representations for each identity, while the attribute compensation network (ACN) generates attribute-wise representations. We explore the complementary relationship between global and attribute-wise representations to maximize intra-domain discrimination. Meanwhile, we design attribute-oriented anti-forgetting (AF) and knowledge consolidation (KC) strategies to minimize inter-domain gaps and facilitate knowledge transfer.}
	\label{fig:fig2}
\end{figure*}
\subsection{Lifelong Person Re-Identification} 
Lifelong Person Re-Identification (LReID) aims to balance intra-domain discrimination with minimizing inter-domain gaps in continuously updated datasets across scenarios, improving the model's anti-forgetting and generalization capabilities. LReID methods can be divided into two categories. 1) \emph{Knowledge distillation-based methods} \cite{huang2023learning, liu2024diverse, yan2024unified, xing2024lifelong} utilize metric strategies to achieve domain-consistent alignment between the old model with learned knowledge distribution and the new model that adaptively learns new knowledge. 2) \emph{Exemplar-based methods} \cite{pu2021lifelong, yu2023lifelong, ge2022lifelong} achieve a distribution balance between old and new samples to prevent catastrophic forgetting by forming a memory buffer to select the limited samples from some identities. These methods strive to reduce inter-domain gaps and ensure consistency across domains to prevent catastrophic forgetting. However, they ignore intra-domain identity discrimination and lack consistency optimization within the inter-domain, limiting the LReID model's performance in learning new knowledge. In this paper, we explore domain consistency representations as a bridge to achieve a trade-off between maximizing intra-domain discrimination and minimizing inter-domain gaps for enhancing the anti-forgetting and generalization capabilities of the LReID model.
\subsection{Vision-Language for Person Re-Identification} 
Vision-language learning paradigms \cite{zhou2022learning, zhou2022conditional} have gained widespread popularity in recent years. Contrastive Language-Image Pre-training (CLIP) \cite{radford2021learning}, establishes a connection between natural language and visual content through the similarity constraint of image-text pairs. CLIP has been applied to multiple person re-identification tasks \cite{yan2023clip, han2023text, qin2024noisy}, including text-to-image, text-based single-modality, and text-based cross-modality. Text-to-image methods \cite{shao2023unified, qin2024noisy, wu2024text} aim to retrieve the target person based on a textual query. Text-based single-modality works \cite{li2023clip, han2023text, yang2024pedestrian} leverage text descriptions to generate robust visual features or integrate the beneficial features of text and images for the person category. Text-based cross-modality methods \cite{du2024yyds} utilize text descriptions to reduce visible-infrared modality gaps. Providing insufficient text descriptions of each identity, due to prompt learning and text inversion. In this paper, we dynamically generate text-image pairs from single images to capture fine-grained global representations based on the CLIP model for improving model performance capability in terms of inter-domain and intra-domain.
\subsection{Pedestrian Attribute Recognition}
Pedestrian attribute recognition aims to assign a set of attributes (Gender, Bag, Short/Long sleeve, and $\textit{etc.}$) to a visual representation of a pedestrian based on their attributes. Deep learning-based research \cite{jia2021rethinking} automatically learns hierarchical features from raw images, improving recognition accuracy. Multi-task learning methods \cite{sun2024adaptive, zhou2024towards, fan2023parformer} leverage additional contextual information from multiple domains, such as pedestrian detection or pose estimation, to significantly improve attribute recognition. Part-based methods \cite{jia2021spatial, jia2022learning} divide the pedestrian image into several parts or regions, providing more accurate localization. Currently, these methods have achieved significant success in improving the accuracy of attribute recognition. We are the first to explore the application of attributes to LReID from two perspectives. 1) Attributes are converted into text descriptions for each image to enhance global representation capabilities. 2) The attributes are transformed into attribute-wise representations by specific networks to maximize intra-domain discrimination and minimize intra-domain gaps.\\
\section{Proposed Method}
\subsection{Preliminary: Overview of Method}
The overview of our DCR to achieve a trade-off between maximizing intra-domain discrimination and inter-domain gaps is shown in Fig. \ref{fig:fig2}. The DCR model learns the old model $\Phi^{t-1}$ and new model $\Phi^t$ from (t-1)-th and t-th steps, where $\Phi^t$ is inherited from $\Phi^{t-1}$. $\Phi^{t-1}$ and $\Phi^t$ with three parts of attribute-text generator (ATG), text-guided aggregation network (TGA), and attribute compensation network (ACN). $\phi^{t-1}$ and $\phi^t$ serve as classifier heads for the old and new models, providing logits of each instance for recognition. Additionally, we define that consecutive $T$ person datasets $D=\{D^t\}_{t=1}^T$ are collected from different environments, and establish a memory buffer $M$ to store a limited number of samples from each previous ReID dataset. Given an image $x_i^t$$\in$$\mathcal{D}^t$$\cup$$\mathcal{M}$, we forward it to $\Phi^{t-1}$ and $\Phi^t$ is as follows:
\begin{equation}
	\begin{aligned}
		&G^{t-1}, AG^{t-1} = \Phi^{t-1}(x_i^t); &G^{t}, AG^{t} = \Phi^{t}(x_i^t). 
	\end{aligned}
\end{equation}
Where $G$ and $AG$ are  global and attribute-wise representations, respectively.
\subsection{Attribute-Text Generator}  
Due to the lack of text-image pairs in ReID datasets, we propose an attribute-text generator (ATG) to dynamically generate text-image pairs for each instance, as illustrated in Fig. \ref{fig:fig2}. Firstly, we employ a pre-trained attribute recognition model \cite{jia2021rethinking} (trained on the PA100K dataset \cite{liu2017hydraplus}) to generate attribute categories. These categories are organized into four major types (e.g., overall, upper body, lower body, and decoration) with 12 subcategories, as shown in TABLE \ref{tab:Table00}. Then, attribute categories are converted into text descriptions for each instance using a predefined template. Finally, the generated text-image pairs are fed into the text-guided aggregation network (TGA) for further processing. In this predefined template (Fig. \ref{fig:fig2}), black font denotes fixed modifiers, while red, blue, orange, and green fonts represent the four major types, respectively. The template dynamically replaces the colored placeholders with the corresponding predicted attributes. To ensure robustness across domains, we set a high confidence threshold (0.80) for attribute classification, thereby maintaining the reliability of text descriptions despite potential domain variations.
\begin{table}[!t]
	\centering
	\renewcommand{\arraystretch}{1.3}
	\setlength{\tabcolsep}{10pt}
	\caption{The attribute categories are divided into four major categories, including 12 subcategories. "No" and "Yes" indicate the absence and presence of this category, respectively.}\label{tab:Table00}
	\begin{tabular}{cccc}  
		\hline
		Type &Category &0 &1\cr
		\hline
		Overall &Sex &man &woman \cr
		\hline
		\multirow{3}{*}{Upper body} &Short sleeved top &No &Yes\cr 
		&Long sleeved top &No &Yes\cr
		&Long coat &No &Yes\cr
		\hline
		\multirow{3}{*}{Lower body} &Trousers &No &Yes\cr
		&Shorts &No &Yes\cr
		&Skirt &No &Yes\cr
		\hline
		\multirow{5}{*}{Decoration}&Hat &No &Yes\cr 
		&Glasses &No &Yes\cr 
		&Handbag &No &Yes\cr 
		&Sholder bag &No &Yes\cr
		&Backpack &No &Yes\cr 
		\hline
	\end{tabular}
\end{table}	
\subsection{Text-Guided Aggregation Network}
To enhance robust representation capabilities for each instance, we propose a text-guided aggregation network (TGA) to explore the global representation for distinguishing identity information in LReID, as shown in Fig. \ref{fig:fig2} (TGA). The TGA consists of a Contrastive Language–Image Pretraining(CLIP) model and a parallel fusion module (PFM). Note that the text encoder is frozen in our DCR model.
\subsubsection{CLIP Model} The CLIP model is a vision-language model that learns multi-modal representations by aligning images and text in a shared embedding space. The model employs a contrastive learning objective to maximize the similarity between an image and its corresponding text description. In our work, we adapt the CLIP model, which includes an image encoder and a text encoder, to generate both image and text embeddings. However, unlike the original CLIP model \cite{radford2021learning}, our model introduces multiple [CLS] embeddings into the image encoder's input sequence. This modification enables the model to capture diverse representations from different views for each instance, thereby enhancing its ability to distinguish fine-grained identity information. Specifically, the clip model generate text embedding $d^*$ and image embedding [$v_1^*$, $\cdots$, $v_N^*$, $v_1$, $\cdots$, $v_P$] for extracting rich text and vision information. image embedding consists of multiple [CLS] embeddings [$v_1^*$, $\cdots$, $v_N^*$] for multi-view representations and patch embedding [$v_1$, $\cdots$, $v_P$] for local feature extraction.
\subsubsection{Parallel Fusion Module} To improve the performance of the LReID model, we propose a parallel fusion module (PFM) to explicitly explore the interactions between image and text embeddings, as shown in Fig. \ref{fig:fig2} (PFM). Firstly, we leverage text embedding $d^*$ as query and image embedding [$v_1^*$, $\cdots$, $v_N^*$, $v_1$, $\cdots$, $v_P$] as key and value to implement operation with cross-attention, drop, and layer normalization, getting text-wise representations. Similarly, in another fusion branch, image-wise representations are obtained. Finally, image-wise and text-wise representations perform concatenation and MLP operations to obtain global representations $G^t=\{G_i^t|i=1,2,\cdots, N\}$, focusing on whole body information. We force multiple global representations $G^t$ at the current step to learn more discriminative information by orthogonal loss to minimize the overlapping elements. The orthogonal loss can be expressed as:
\begin{equation}
	L_{Ort} = \sum\limits_{i=1}^{N-1}\sum\limits_{j=i+1}^{N}(G^t_i,G^t_j).
\end{equation}
Then, we utilize the cross-entropy loss $L_{\mathrm{CE}}$ and triplet loss $L^g_{\mathrm{Tri}}$ \cite{yang2024pedestrian} to optimize our DCR at the current task.
\begin{equation}
	L_{\mathrm{CE}}=\frac{1}{K}\sum\limits_{i=1}^{K}y_i\log((\phi^t(G^t))_i).
\end{equation}
\begin{equation}
	L^g_{\mathrm{Tri}} = max(d^g_p-d^g_n+m, 0).
\end{equation}
Where $K$ is the number of classes, and $m$ is the margin, $d^g_p$ and $d^g_n$ are the distances from positive samples and negative samples to anchor samples in global representations, respectively. Unlike some methods \cite{wu2021generalising, yu2023lifelong}, global representations generated by the text-guided aggregation (TGA) network present two advantages. First, we leverage text descriptions based on the CLIP model to enhance the discrimination capability of global representations, allowing them to better distinguish identities and adapt to new knowledge. Second, global representations facilitate knowledge transfer, improving the model's generalization ability.
\subsection{Attribute Compensation Network} 
We force attributes to guide the attribute compensation network (ACN) for learning attribute-wise representations. The ACN consists of an attribute decoder and an attribute matching component, as illustrated in Fig. \ref{fig:fig2} (ACN).
\subsubsection{Attribute Decoder} Enabling attributes to better adapt across domains, we define multiple learnable attributes semantic information $S^t=\{S^t_i|i=1,2,\cdots, N\}$ to learn discriminative information. The attributes undergo a linear layer to increase their dimensions and are then multiplied with the text-image global representation to output $f_{AT}$. Attribute semantic information $S^t$ as queries $Q$, $f_{AT}$ as keys and values are input into the attribute decoder, which outputs the attribute features $A^t=\{A_i^t|i=1,2,\cdots, N\}$. The attribute decoder utilizes six transformer blocks (T$\_$Block) referenced from \cite{vaswani2017attention}.  
\subsubsection{Attribute Matching} The attribute features $A^t=\{A_i^t|i=1,2,\cdots, N\}$ focus on multiple discriminative local details of individuals. However, it remains unclear which views make these attribute features more prominent in the global representations. We propose an Attribute Matching (AM) mechanism to select the optimal alignment between attribute features and global representations, thereby identifying the best view for the global representation and improving the representational ability of the attribute features. Specifically, we first calculate the similarity between the attribute features $A^t=\{A_i^t|i=1,2,\cdots, N\}$ and global representations $G^t=\{G_i^t|i=1,2,\cdots, N\}$ from different views. Based on this matching process, we determine the index of the most similar global representation as follows.
\begin{equation}
	k = argmax(<A_i^t,G^t>).
\end{equation}
where $<,>$ represent cosine similarity. The $argmax$ operation selects both the maximum similarity value and its corresponding index. We identify the most similar index between each attribute feature $A_i|i=1,2,\cdots, N$ and the global representations  $G$, and then combine the attribute feature $A_i$ and with its corresponding global representation $G_k^t$ (where $k$ is the matched index) to obtain the attribute-wise representations  $AG^t=\{AG_i|i=1,2,\cdots,N\}$. This process is defined as:
\begin{equation}
	AG_i^t = A_i^t + G_k^t.
\end{equation}
\indent We leverage the triplet loss to align attribute-wise representations with identity at the current step, assisting in global representations to distinguish similar identities. 
\begin{equation}
	L^l_{\mathrm{Tri}} = max(d_p-d_n+m, 0).
\end{equation}
where, $d^l_p$ and $d^l_n$ are the distances from positive samples and negative samples to anchor samples in attribute-wise representations, respectively. In this paper, attribute-wise representations that contain specific information of individuals assist global representations in distinguishing similar identities for maximizing intra-domain discrimination. Meanwhile, attribute-wise representations as a bridge across increasing datasets to minimize inter-domain gaps for better knowledge transfer.
\subsection{Attribute-oriented Anti-Forgetting}
We develop an attribute-oriented anti-forgetting (AF) strategy to explore attribute-wise representations that align the distributions of the old and new models, as shown in Fig. \ref{fig:fig2} (AF). The new model can adapt to new information but may forget old knowledge from the previous datasets, while the old model retains old knowledge. To preserve old knowledge, we leverage attribute-wise representations as a bridge to optimize both the old and new models by using samples from the memory buffer. This strategy achieves domain consistency and minimizes inter-domain gaps, alleviating the forgetting of old knowledge, and is calculated as follows:
\begin{equation}
	L_{AF} = \frac{1}{B}\sum\limits_{i=1}^{B}KL(AG^{t-1}_N/\tau||AG^t_N/\tau).
\end{equation}
Where $KL(.||.)$ is a kullback-leibler divergence, and $\tau$ represents a hyper-parameter called temperature \cite{hinton2015distilling}.
\subsection{Knowledge Consolidation}
Maximizing intra-domain discrimination and minimizing inter-domain gaps are in a contradictory relationship. Therefore, achieving a balance between them is crucial for improving the performance of LReID models. Thus, we propose a knowledge consolidation (KC) strategy that leverages global representations for knowledge transfer between old and new models. This includes alignment mechanism and logit-level distillation mechanism.\\
\indent Maintaining distribution consistency between the old and new models for previous datasets can limit the model's ability to learn new knowledge. Therefore, we propose an alignment mechanism to explore global representations of knowledge transfer from the current dataset, as follows:
\begin{equation}
	L_{AL} = \frac{1}{B}\sum\limits_{i=1}^{B}KL(G^{t-1}/\tau||G^t/\tau).
\end{equation}
\indent We further introduce a logit-level distillation mechanism to enhance the extraction of identity information shared between the old and new models, further improving the model's knowledge consolidation ability. This is represented as follows:
\begin{equation}
	L_{LD} = \frac{1}{B}\sum\limits_{i=1}^{B}KL((\phi^{t-1}(G^{t-1}))_i/\tau||(\phi^t(G^{t}))_i/\tau).
\end{equation}
\indent The knowledge consolidation loss is defined as:
\begin{equation}
	L_{KC} = L_{AL} + L_{LD}.
\end{equation}
\indent The total loss function is formulated as:
\begin{equation}
	L = L_{CE} + L^g_{Tri} + L^l_{Tri} + L_{Ort} + L_{AF} + L_{KC}.
\end{equation}
\begin{table}[htbp]
	\centering
	\renewcommand{\arraystretch}{1.3}
	\setlength{\tabcolsep}{7pt}
	\caption{Dataset statistics for both seen and unseen domains. Since the selection process resulted in 500 train IDs being selected, the original numbers of IDs are listed for comparison. '-' denotes that the dataset is not used for training.}\label{tab:Table0}
	\begin{tabular}{l|l|l|c|c}  
		\hline
		Type &Datasets &Scale &Train IDs &Test IDs \cr
		\hline
		\multirow{5}{*}{Seen}&
		Market1501 \cite{zheng2015scalable} &large &500(751) &750 \cr
		&CUHK-SYSU \cite{xiao2016end} &mid &500(942) &2900 \cr
		&DukeMTMC \cite{ristani2016performance} &large &500(702) &1110 \cr
		&MSMT17$\_$V2 \cite{wei2018person} &large &500(1041) &3060 \cr
		&CUHK03 \cite{li2014deepreid} &mid &500(700) &700 \cr
		\hline
		\multirow{6}{*}{Unseen}&
		VIPeR \cite{gray2008viewpoint} &small &\makecell[c]{$-$} &316  \cr
		&GRID \cite{loy2010time} &small &\makecell[c]{$-$} &126 \cr
		&CUHK02 \cite{li2013locally} &mid &\makecell[c]{$-$} &239 \cr
		&Occ$\_$Duke \cite{miao2019pose} &large &\makecell[c]{$-$} &1100 \cr
		&Occ$\_$REID \cite{zhuo2018occluded} &mid &\makecell[c]{$-$} &200 \cr
		&PRID2011 \cite{hirzer2011person} &small &\makecell[c]{$-$} &649 \cr
		\hline
	\end{tabular}
\end{table}
\begin{table*}[ht]
	\centering
	\renewcommand{\arraystretch}{1.3}
	\setlength{\tabcolsep}{8pt}	
	\caption{Performance comparison with state-of-the-art methods on training order-1. Bold and red fonts are optimal and suboptimal values, respectively. Training order-1 is Market1501$\to$CUHK-SYSU$\to$ DukeMTMC$\to$MSMT17$\_$V2$\to$CUHK03.}\label{tab:Table1}
	\begin{tabular}{c|cc|cc|cc|cc|cc|cc|cc}
		\hline
		\multirow{2}{*}{Method}&
		\multicolumn{2}{c|}{Market1501}&\multicolumn{2}{c|}{CUHK-SYSU}&\multicolumn{2}{c|}{DukeMTMC}&\multicolumn{2}{c|}{MSMT17$\_$V2}&
		\multicolumn{2}{c|}{CUHK03}&\multicolumn{2}{c}{Seen-Avg}&\multicolumn{2}{c}{Unseen-Avg}\\
		\cline{2-15}
		&mAP&R-1&mAP&R-1&mAP&R-1&mAP&R-1&mAP&R-1 &mAP &R-1 &mAP &R-1\cr
		\hline
		AKA\cite{pu2021lifelong} &58.1 &77.4 &72.5 &74.8 &28.7 &45.2 &6.1 &16.2 &38.7 &40.4 &40.8 &50.8 &42.0 &39.8\\
		PTKP\cite{ge2022lifelong} &64.4 &82.8 &79.8 &81.9 &45.6 &63.4 &10.4 &25.9 &42.5 &42.9 &48.5 &59.4 &51.2 &49.1 \\
		PatchKD\cite{sun2022patch} &68.5 &85.7 &75.6 &78.6 &33.8 &50.4 &6.5 &17.0 &34.1 &36.8 &43.7 &53.7 &45.1 &43.3\\
		KRKC\cite{yu2023lifelong} &54.0 &77.7 &83.4 &\textcolor{red}{85.4} &48.9 & 65.5 &14.1 &33.7 &49.9 &50.4 &50.1 &62.5 &\textcolor{red}{52.7} &\textcolor{red}{50.8} \\
		ConRFL\cite{huang2023learning} &59.2 &78.3 &82.1 &84.3 &45.6 &61.8 &12.6 &30.4 &\textcolor{red}{51.7} &\textcolor{red}{53.8} &50.2 &61.7 &- &-\\
		CODA\cite{smith2023coda} &53.6 &76.9 &75.7 &78.1 &48.6 &59.5 &13.2 &31.3 &47.2 &48.6 &47.7 &58.9 &44.5 &42.4\\
		LSTKC\cite{xu2024lstkc} &54.7 &76.0 &81.1 &83.4 &49.4 &66.2 &20.0 &43.2 &44.7 &46.5 &50.0 &63.1 &51.3 &48.9\\
		C2R\cite{cui2024learning} &\textcolor{red}{69.0} &\textcolor{red}{86.8} &76.7 &79.5 &33.2 &48.6 &6.6 &17.4 &35.6 &36.2 &44.2 &53.7 &- &-\\
		DKP\cite{xu2024distribution} &60.3 &80.6 &\textcolor{red}{83.6} &\textcolor{red}{85.4} &\textcolor{red}{51.6} &\textcolor{red}{68.4} &\textcolor{red}{19.7} &\textcolor{red}{41.8} &43.6 &44.2 &\textcolor{red}{51.8} &\textcolor{red}{64.1} &49.9 &46.4\\
		\hline
		Baseline &61.6 &79.1 &80.2 &80.6 &50.2 &64.3 &15.1 &36.5 &44.9 &46.8 &50.4 &61.5 &51.8 &49.4\\
		Ours &\textbf{75.9} &\textbf{87.9} &\textbf{87.3} &\textbf{88.5} &\textbf{60.1} &\textbf{71.9} &\textbf{25.3} &\textbf{50.1} &\textbf{60.5} &\textbf{61.3} &\textbf{61.8} &\textbf{71.9} &\textbf{60.8} &\textbf{58.3}\\							
		\hline
	\end{tabular}
\end{table*}
\begin{table*}[!ht]
	\centering
	\renewcommand{\arraystretch}{1.3}
	\setlength{\tabcolsep}{8pt}
	\caption{Performance comparison with state-of-the-art methods on training order-2. Bold and red fonts are optimal and suboptimal values, respectively. Training order-2 is DukeMTMC$\to$MSMT17$\_$V2$\to$Market1501$\to$ CUHK-SYSU$\to$CUHK03.}\label{tab:Table2}
	\begin{tabular}{c|cc|cc|cc|cc|cc|cc|cc}
		\hline
		\multirow{2}{*}{Method}&
		\multicolumn{2}{c|}{DukeMTMC}&\multicolumn{2}{c|}{MSMT17$\_$V2}&\multicolumn{2}{c|}{Market1501}&\multicolumn{2}{c|}{CUHK-SYSU}& \multicolumn{2}{c|}{CUHK03}&\multicolumn{2}{c}{Seen-Avg}&\multicolumn{2}{c}{Unseen-Avg}\\
		\cline{2-15}
		&mAP &R-1 &mAP &R-1 &mAP &R-1 &mAP &R-1 &mAP &R-1 &mAP &R-1 &mAP &R-1 \cr 
		\hline
		AKA\cite{pu2021lifelong} &42.2 &60.1 &5.4 &15.1 &37.2 &59.8 &71.2 &73.9 &36.9 &37.9 &38.6 &49.4 &41.3 &39.0 \\
		PTKP\cite{ge2022lifelong} &54.8 &70.2 &10.3 &23.3 &59.4 &79.6 &80.9 &82.8 &41.6 &42.9 &49.4 &59.8 &50.8 &48.2\\
		PatchKD\cite{sun2022patch} &58.3 &74.1 &6.4 &17.4 &43.2 &67.4 &74.5 &76.9 &33.7 &34.8 &43.2 &54.1 &44.8 &43.3\\
		KRKC\cite{yu2023lifelong} &50.6 &65.6 &13.6 &27.4 &56.2 &77.4 &\textcolor{red}{83.5} &\textcolor{red}{85.9} &46.7 &46.6 &50.1 &61.0 &52.1 &47.7 \\
		ConRFL\cite{huang2023learning} &34.4 &51.3 &7.6 &20.1 &\textcolor{red}{61.6} &80.4 &82.8 &85.1 &\textcolor{red}{49.0} &\textcolor{red}{50.1} &47.1 &57.4 &- &-\\
		CODA\cite{smith2023coda} &38.7 &56.6 &11.6 &24.5 &54.3 &75.1 &76.2 &75.8 &42.3 &41.7 &44.6 &54.7 &45.0 &42.9\\
		LSTKC\cite{xu2024lstkc} &49.9 &67.6 &\textcolor{red}{14.6} &\textcolor{red}{34.0} &55.1 &76.7 &82.3 &83.8 &46.3 &48.1 &49.6 &62.1 &51.7 &49.5\\		
		C2R\cite{cui2024learning} &\textcolor{red}{59.7} &\textcolor{red}{75.0} &7.3 &19.2 &42.4 &66.5 &76.0 &77.8 &37.8 &39.3 &44.7 &55.6 &- &-\\
		DKP\cite{xu2024distribution} &53.4 &70.5 &14.5 &33.3 &60.6 &\textcolor{red}{81.0} &83.0 &84.9 &45.0 &46.1 &\textcolor{red}{51.3} &\textcolor{red}{63.2} &51.3 &47.8\\
		\hline
		Baseline &53.8 &69.1 &14.1 &29.8 &59.8 &80.4 &78.4 &78.5 &45.3 &44.9 &50.3 &60.5 &\textcolor{red}{52.2} &\textcolor{red}{49.9}\\
		Ours &\textbf{64.1}&\textbf{77.2}&\textbf{25.4} &\textbf{44.9} &\textbf{70.6} &\textbf{84.5} &\textbf{86.1} &\textbf{88.2} &\textbf{54.2} &\textbf{58.7} &\textbf{60.1} &\textbf{70.7} &\textbf{61.6} &\textbf{59.2} \\
		\hline
	\end{tabular}
	
\end{table*}
\section{Experiments}
\subsection{Experiments Setting}
\subsubsection{Datasets} To assess the performance of our method in anti-forgetting and generalization, we evaluate our method on a challenging benchmark consisting of Market1501 \cite{zheng2015scalable}, CUHK-SYSU \cite{xiao2016end}, DukeMTMC \cite{ristani2016performance}, MSMT17$\_$V2 \cite{wei2018person} and CUHK03 \cite{li2014deepreid}, referred to as the seen domain. Two representative training orders are set up following the protocol described in \cite{pu2021lifelong} for training and testing. Further, we employ six datasets including VIPeR \cite{gray2008viewpoint}, GRID \cite{loy2010time},  CUHK02 \cite{li2013locally}, Occ$\_$Duke \cite{miao2019pose}, Occ$\_$REID \cite{zhuo2018occluded}, and PRID2011 \cite{hirzer2011person}, as the unseen domain. During evaluation, Unseen domain and test sets of the seen domain are combined into a single benchmark. Detailed statistics for these datasets can be shown in TABLE \ref{tab:Table0}.
\subsubsection{Implementation Details} Our text encoder and image encoder are based on a pre-trained CLIP model, while the attribute decoder utilizes a transformer-based architecture\cite{vaswani2017attention}. All person images are resized to 256$\times$128. We use Adam \cite{kingma2014adam} for optimization and train each task for 60 epochs. The batch size is set to 128. The learning rate is initialized at 5$\times$$10^{-6}$ and is decreased by a factor of 0.1 every 20 epochs for each task. We employ mean average precision (mAP) and Rank-1 accuracy (R-1) to evaluate the LReID model on each dataset.
\subsection{Comparison with SOTA Methods}
We compare the proposed DCR with SOTA LReID to demonstrate the superiority of our method, including AKA\cite{pu2021lifelong}, PTKP\cite{ge2022lifelong}, PatchKD\cite{sun2022patch}, KRKC\cite{yu2023lifelong}, ConRFL\cite{huang2023learning}, CODA \cite{smith2023coda}, LSTKC\cite{xu2024lstkc}, C2R\cite{cui2024learning}, DKP\cite{xu2024distribution}. Experimental results on training order-1 and order-2 are shown in TABLE \ref{tab:Table1} and TABLE \ref{tab:Table2}, respectively.
\subsubsection{Compared with LReID methods on the Seen Domain} In TABLE \ref{tab:Table1} and TABLE \ref{tab:Table2}, our DCR significantly outperforms LReID methods, with an seen-avg incremental gain of  10.0\% mAP/7.8\% R-1, and 9.8\% mAP/7.5\% R-1 on training order-1 and order-2, respectively. Meanwhile, our DCR effectively alleviates catastrophic forgetting, achieving 6.9\% mAp/1.1\% R-1 and 5.4\% mAP/2.2\% R-1 improvement on the first dataset (Mrket1501 and DukeMTMC) with different training orders. Compared to CODA, our DCR significantly outperforms performance under the backbone of VIT-B/16. In contrast, our DCR achieves a trade-off between anti-forgetting and acquiring new information.
\subsubsection{Compared with LReID methods on the Unseen Domain} In Table \ref{tab:Table1}, compared to the KRKC method, our DCR model achieves significant improvements, with an average gain of 8.1\% mAP / 7.5\% R-1 on the unseen domain. Furthermore, when evaluated against the LSTKC methods, our DCR demonstrates superior performance over LReID methods, delivering an unseen-avg incremental improvement of 9.5\% mAP/11.0\% R-1, as shown in Table \ref{tab:Table2}. We attribute these advancements to the complementary relationship between global and attribute-wise representations, which effectively enhances discrimination among similar identities. Additionally, our attribute-oriented anti-forgetting (AF) and knowledge consolidation (KC) strategies foster inter-domain consistency and facilitate seamless knowledge transfer. In summary, our DCR markedly strengthens generalization capabilities, outperforming existing approaches in terms of mAP and R-1 metrics.
\subsubsection{Compared with Baseline} Due to the lack of CLIP-based comparison methods in LReID, we introduce a Baseline model that includes the CLIP model, an attribute-text generator, and a knowledge consolidation strategy. The Baseline outperforms other methods (such as AKA, PTKP, PatchKD, $\textit{etc.}$) in mAP and R-1, benefiting from the powerful extraction capabilities of CLIP, as presented in TABLE \ref{tab:Table1} and TABLE \ref{tab:Table2}. Compared to the Baseline, our DCR improves the Seen-Avg by 11.4\% mAP/10.4\% R-1 and by 9.8\% mAP/10.2\% R-1. These results demonstrate that our proposed domain consistency representation learning strategy achieves significant performance in balancing the maximization of intra-domain discrimination and the minimization of inter-domain gaps in LReID.    
\begin{figure}[t]
	\centering 
	\includegraphics[width=0.96\linewidth, height=0.16\textheight]{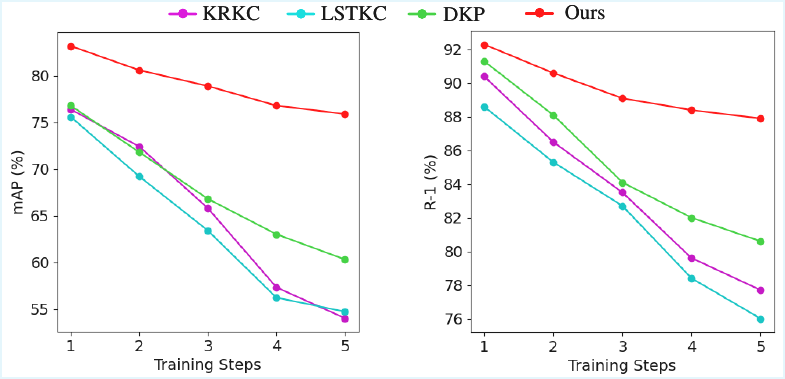}
	\caption{Anti-forgetting curves. After each training step, we measure the metrics of Market1501 in the training order-1 to demonstrate the model's anti-forgetting performance.}
	\label{fig:fig33}
\end{figure}
\subsubsection{The Anti-forgetting Performance of Our Method} We conduct a forgetting measurement experiment in training order-1, as shown in Fig. \ref{fig:fig33}. The Fig. \ref{fig:fig33} shows the metric measurements for the Market1501 dataset at different training steps. After training on the large-scale MSTMS17 dataset at training step 4, KRKC, LSTCC, and DKP exhibit significant attenuation in mAP and R-1. Because the comparison method limits the performance of the model in minimizing inter-domain gaps. Our method demonstrates a smoother decrease in mAP and Rank-1, which can effectively reduce inter-domain gaps to alleviate the catastrophic forgetting problem.
\begin{figure*}[t]
	\centering 
	\includegraphics[width=1.0\linewidth, height=0.20\textheight]{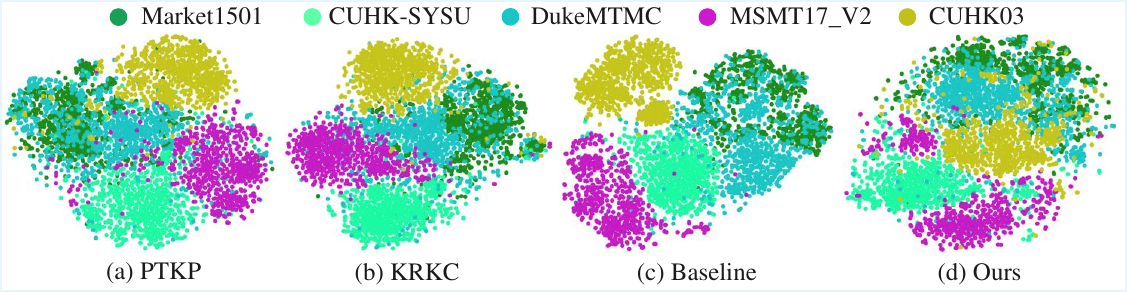}
	\caption{t-SNE visualization of feature distribution on five datasets from the seen domain. Our DCR effectively improves the anti-forgetting and generalization ability, which significantly distinguishes identity information within a domain but also spreads identity information across multiple domains.}
	\label{fig:fig3}
\end{figure*}
\begin{figure}[t]
	\centering 
	\includegraphics[width=1\linewidth, height=0.16\textheight]{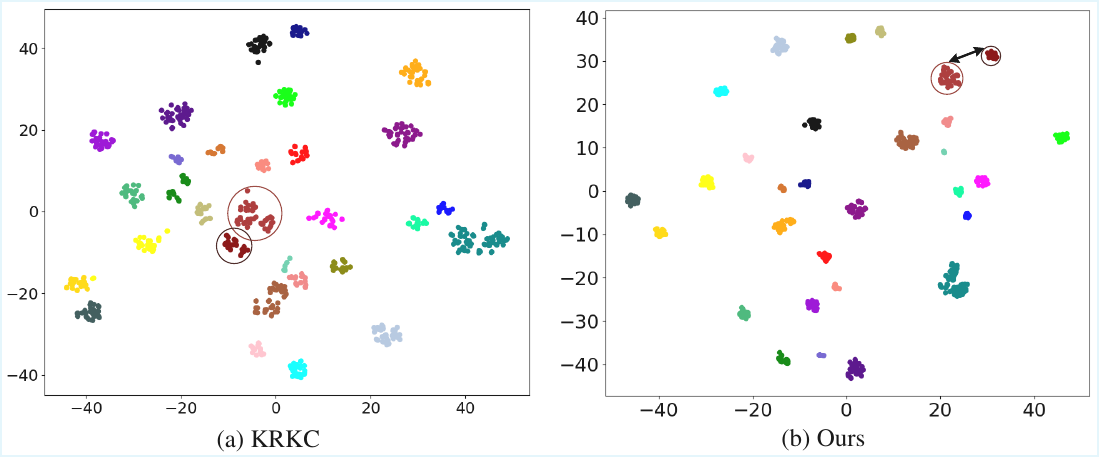}
	\caption{Visualization of intra-domain discrimination on the Market1501 dataset. We randomly select 30 identities. Colors represent different identity information. Our DCR can cluster images of the same identity more tightly (circle) for minimizing inter-domian discrimination.}
	\label{fig:fig4}
\end{figure}
\subsubsection{The effectiveness of minimizing inter-domain gaps} We visualize the feature distribution of PTKP, KRKC, Baseline, and our method across five datasets as shown in Fig. \ref{fig:fig3}. The Baseline shows poor performance in bridging inter-domain gaps, as the lack of attribute-wise representations makes it challenging to reduce inter-domain gaps. The KRKC method effectively separates each domain, but it insufficiently distinguishes identity information within the domain, limiting the model's ability to prevent forgetting and enhance generalization. Compared to other methods, our DCR effectively distinguishes identity information within a domain and spreads identity information across multiple domains, which significantly improves the anti-forgetting and generalization ability of the model.
\subsubsection{The effectiveness of maximizing intra-domain discrimination} We visualize the feature distribution of KRKC and our method. Fig. \ref{fig:fig4} shows that our DCR can significantly cluster images of the same identity more tightly (circle) and increase the distance between different identities (black bidirectional arrow). Compared to KRKC, our DCR improves intra-domain discrimination due to the complementary relationship between global and attribute-wise representations, which enables it to learn the subtle nuances of individuals.\\
\begin{figure}[t]
	\centering 
	\includegraphics[width=0.96\linewidth, height=0.18\textheight]{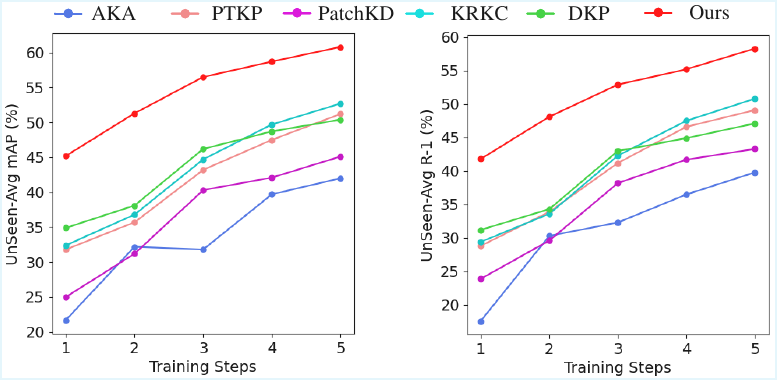}
	\caption{Generalization curves. After each training step, the performance of the unseen domain is evaluated.}
	\label{fig:fig5}
\end{figure}
\subsubsection{Generalization Curves on the Unseen Domain} We analyze the average performance on the unseen domain during the training steps, as depicted in Fig. \ref{fig:fig5}. Compared to other methods, our DCR achieves superior performance and exhibits faster performance growth across the training steps. Thus, our attribute-oriented anti-forgetting (AF) strategy effectively bridges inter-domain gaps and enhances the generalization ability of our model. In summary, our DCR explores global and attribute-wise representations to achieve a trade-off between maximizing intra-domain discrimination and minimizing inter-domain gaps.
\begin{table}[!ht]
	\centering
	\renewcommand{\arraystretch}{1.3}
	\setlength{\tabcolsep}{8pt}
	\caption{Ablation studies on the number of global and attribute-wise representations $N$ on training order-1.}
	\begin{tabular}{c|c|c|c|c}
		\hline
		\multirow{2}{*}{Number ($N$)}& \multicolumn{2}{c|}{Seen$\_$Avg}&\multicolumn{2}{c}{Unseen$\_$Avg}\\
		\cline{2-5}
		&mAP &R-1 &mAP &R-1 \cr \hline
		2 &60.2 &68.7 &59.4 &56.5  \\
		3 &\textbf{61.8}&\textbf{71.9}&\textbf{60.8}&\textbf{58.3}  \\ 
		4 &61.2 &71.6 &60.3 &57.5  \\
		\hline
	\end{tabular}
	\label{tab:Table3}
\end{table}
\begin{table}[!ht]
	\centering
	\renewcommand{\arraystretch}{1.3}
	\setlength{\tabcolsep}{6pt}
	\caption{Ablation studies of different components on training order-1.}\label{tab:Table4}
	\begin{tabular}{c|cccc|c|c|c|c}
		\hline
		\multirow{2}{*}{Type} &\multicolumn{4}{c|}{Components} &\multicolumn{2}{c|}{Seen$\_$Avg} &\multicolumn{2}{c}{Unseen$\_$Avg}\\
		\cline{2-9}
		&PFM &ACN &AF &KC &mAP &R-1 &mAP &R-1 \cr \hline
		(a)&&&&&50.4&61.5&51.8&49.4  \\
		(b) &$\surd$ &&&&51.7 &62.1 &52.5 &50.3 \\
		(c) &$\surd$ &$\surd$ &&&56.9 &68.3 &57.6 &55.4 \\
		(d) &$\surd$ &$\surd$ &$\surd$ &&58.7 &69.2 &58.5 &56.8  \\
		(e) &$\surd$ &$\surd$ &$\surd$ &$\surd$ &\textbf{61.8} &\textbf{71.9} &\textbf{60.8} &\textbf{58.3}  \\ 
		\hline
	\end{tabular}
\end{table}\\
\subsection{Ablation Studies}
\subsubsection{The number of global and attribute-wise representations} Global and attribute-wise representations capture individual nuances in intra-domain and inter-domain consistency. We evaluate the suitability of multiple global and attribute-wise representations as shown in TABLE \ref{tab:Table3}. We have observed that setting the number of global and attribute-wise representations $N$ to 3 achieves the best performance for our method.
\subsubsection{Performance of Different Components} To evaluate the contribution of each component to our DCR model, we conduct ablation studies on both seen and unseen domains, as detailed in Table \ref{tab:Table4}. Here, PFM denotes the Parallel Fusion Module, while ACN refers to the Attribute Compensation Network. AF and KC represent the attribute-oriented anti-forgetting and knowledge consolidation strategies, respectively. Comparing (a) and (b), we observe that PFM improves performance by +1.3\% mAP / +0.6\% R-1 on seen$\_$avg and +0.7\% mAP / +0.9\% R-1 on unseen$\_$avg under the baseline method. This demonstrates its effectiveness in fusing text and image information to generate diverse global representations. Comparing (a) and (c), integrating PFM and ACN yields significant gains: +6.5\% mAP / +6.8\% R-1 on seen$\_$avg and +5.8\% mAP / +6.0\% R-1 on unseen$\_$avg, for generating the diverse global representations. Comparing (c) and (e), when integrating AF and KC, the model achieves gains of +4.9\% mAP / +3.6\% R-1 on seen$\_$avg and +3.2\% mAP / +2.9\% R-1 on unseen$\_$avg. These strategies significantly promote inter-domain consistency and knowledge transfer. Our DRE method unifies PFM, ACN, AF, and KC into an end-to-end LReID model, striking a balance between maximizing intra-domain discrimination and minimizing inter-domain gaps. These combinations ensure strong generalization while mitigating catastrophic forgetting in our DCR.
\begin{figure*}[t]
	\centering 
	\includegraphics[width=1.0\linewidth, height=0.33\textheight]{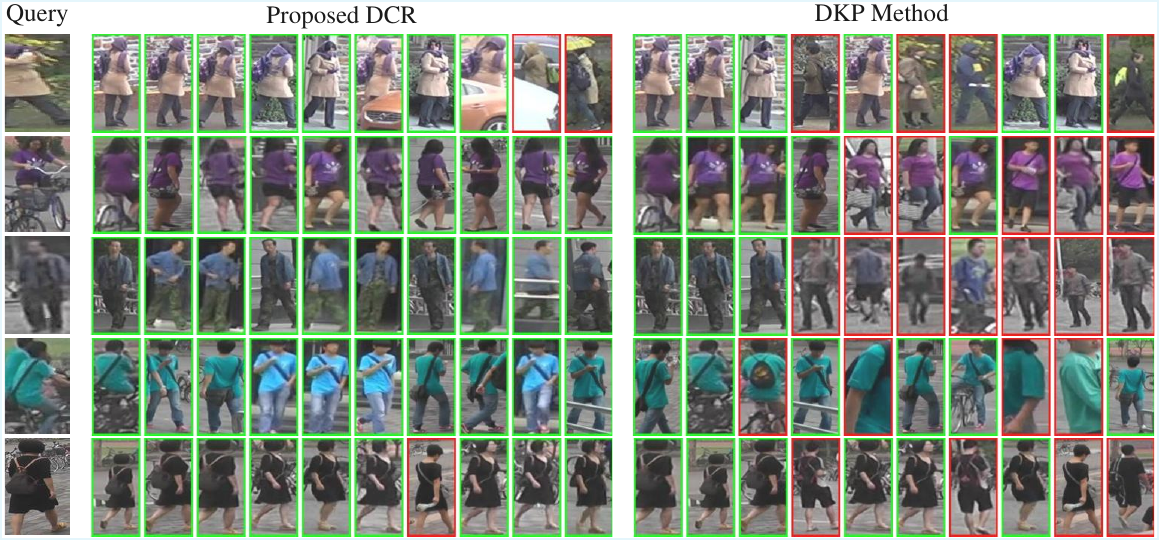}
	\caption{Visualization of retrieval results. The left column displays the query image under challenging conditions, such as occlusion, blur, side view, and back view. In the retrieved results, correctly matched pedestrians are marked with green boxes, while incorrect matches are highlighted in red box. Compared to the DKP method, our DCR retrieves more pedestrians with the same identity, demonstrating superior performance.}
	\label{fig:fig6}
\end{figure*}
\begin{table}[!ht]
	\centering
	\renewcommand{\arraystretch}{1.3}
	\setlength{\tabcolsep}{8pt}
	\caption{Ablation of training with or without attribute-text generator (ATG) on training order-1.}
	\begin{tabular}{c|c|c|c|c}
		\hline
		\multirow{2}{*}{Method}& \multicolumn{2}{c|}{Seen$\_$Avg}&\multicolumn{2}{c}{Unseen$\_$Avg}\\
		\cline{2-5}
		&mAP &R-1 &mAP &R-1 \cr \hline
		Training w/o ATG &60.1 &70.5 &59.3 &56.5  \\
		Training w/ ATG &\textbf{61.8}&\textbf{71.9}&\textbf{60.8}&\textbf{58.3}  \\ 
		\hline
	\end{tabular}
	\label{tab:Table5}
\end{table}\\
\begin{table}[!ht]
	\centering
	\renewcommand{\arraystretch}{1.3}
	\setlength{\tabcolsep}{8pt}
	\caption{Ablation studies of confidence threshold on training order-1.}
	\begin{tabular}{c|c|c|c|c}
		\hline
		\multirow{2}{*}{Confidence Threshold} &\multicolumn{2}{c|}{Seen$\_$Avg} &\multicolumn{2}{c}{Unseen$\_$Avg}\\
		\cline{2-5}
		&mAP &R-1 &mAP &R-1 \cr \hline
		0.7 &60.9 &71.1 &59.7 &57.6  \\
		0.8 &61.8 &\textbf{71.9} &\textbf{60.8} &\textbf{58.3}  \\ 
		0.9 &\textbf{62.0} &71.6 &60.5 &58.0  \\
		\hline
	\end{tabular}	
	\label{tab:Table6}
\end{table}\\
\subsubsection{Performance of attribute-text generator} To better understand whether each instance's text descriptions generated by the attribute-text generator (ATG) provide more fine-grained guidance for learning global representations, we train our model using the generic text descriptor "A photo of a person" (w/o ATG) for comparison. TABLE \ref{tab:Table5} shows that the attribute-text generator obtains text descriptions to significantly improve overall performance. When using the specific text descriptors, the average decreases by 1.7\% mAP/1.4\% R-1 on the seen domain and by 1.5\% mAP/1.8\% R-1 on the unseen domain. ATG enhances the robustness of global representations for each instance, effectively mitigating the forgetting of old knowledge.
\subsubsection{Analysis of confidence threshold} To evaluate the sensitivity of the confidence threshold in an attribute-text generation, we conduct an ablation study with threshold values of 0.7, 0.8, and 0.9, as shown in TABLE \ref{tab:Table6}. The results demonstrate that increasing the confidence threshold from 0.7 to 0.8 consistently enhances DCR performance across metrics. While a confidence threshold of 0.9 achieves the highest 62.0\% mAP on seen$\_$avg, other metrics do not reach their peak performance at this value. Overall, we find that setting the confidence threshold to 0.8 strikes the best balance, ensuring robust classification accuracy while maximizing model performance.
\subsection{Visualization} 
To further validate the effectiveness of our DCR, we conduct a qualitative comparison with the DKP method, as shown in Fig. \ref{fig:fig6}. The left column displays challenging query images, including cases with occlusion, blur, side views, and back views. In Fig. \ref{fig:fig6}, correctly matched pedestrians are marked with green boxes, while incorrect matches are highlighted in red. Under occlusion conditions (first and second rows), our DCR accurately retrieves eight out of ten individuals with the same identity, despite the limited visual information available. For blurry queries (third row), the DKP method only correctly matches three pedestrians, whereas our DCR achieves perfect retrieval (10/10) even with the unclear query image. Additionally, in side-view and back-view scenarios (fourth and fifth rows), DCR successfully handles clothing color variations, outperforming DKP in matching the same identity.	These results demonstrate that DCR achieves superior retrieval performance, consistently identifying more pedestrians with the same identity across diverse challenging conditions. This advantage stems from the collaborative design of global and attribute-wise representations, which balances intra-domain discrimination and inter-domain gaps. By enhancing discrimination between similar identities and adapting to complex scenarios, our DCR demonstrates significantly more robustness than existing approaches.
\begin{table}[t]
	\centering
	\renewcommand{\arraystretch}{1.3}
	\setlength{\tabcolsep}{6pt}
	\caption{Experimental results regarding training time and inference time on the Market1501.}
	\label{tab:Table8}
	\begin{tabular}{c|c|c}
		\hline
		Method & Training time (h-min) & inference time (s)\cr \hline
		AKA\cite{pu2021lifelong} & 4h-50min &25s  \\
		PTKP\cite{ge2022lifelong} &16h-19min &20s \\
		PatchKD\cite{sun2022patch} &3h-30min &103s  \\ 
		KRKC\cite{huang2023learning} &17h-31min &56s  \\
		DKP\cite{xu2024distribution} &3h-50min &38s  \\
		Ours &19h-10min &65s  \\
		\hline  
	\end{tabular}
\end{table}
\subsection{Analysis of Speed} 
We evaluate the computational efficiency and performance of various methods on the Market1501 dataset, which consists of a query set (750 IDs, 3,368 images) and a gallery set (751 IDs, 15,913 images). All experiments were conducted using two NVIDIA RTX A6000 GPUs for training and a single GPU for inference. As shown in Table \ref{tab:Table8}, where time measurements are indicated in hours (h), minutes (min), and seconds (s), our DCR completes training in 19 hours and 10 minutes while achieving an inference time of 65 seconds, processing gallery images at a rate of 52 images per second. Although our DCR shows longer training and inference times compared to AKA, PatchKD, and DKP approaches, it delivers significantly superior re-identification performance, achieving an seen$\_$avg incremental gain of 10.0\% mAP/7.8\% R-1, and 9.8\% mAP/7.5\% R-1, an seen$\_$avg improvement 10.0\% mAP/7.8\% R-1, and 9.8\% mAP/7.5\% R-1 on training order-1 and order-2, as demonstrated in Tables \ref{tab:Table1} and \ref{tab:Table2}. The increased computational complexity of DCR stems from its transformer-based CLIP architecture incorporating multiple class tokens and the additional attribute compensation network (ACN) that enhances discrimination of similar identities through local feature integration. While this design choice results in higher computational demands, it achieves an excellent balance between performance metrics and complexity. For future work, we plan to explore prompt tuning \cite{smith2023coda} and mixture-of-experts (MOE) approaches \cite{yu2024boosting} to optimize the model's efficiency while maintaining its competitive performance advantages.

\section{Conclusions}
In this paper, we propose a domain consistency representation learning (DCR) model that explores global and attribute-wise representations to capture subtle nuances in intra-domain and inter-domain consistency, achieving a trade-off between maximizing intra-domain discrimination and minimizing inter-domain gaps. Specifically, global and attribute-wise representations serve as complementary information to distinguish similar identities within the domain. We also develop an attribute-oriented anti-forgetting (AF) strategy and a knowledge consolidation (KC) strategy to minimize inter-domain gaps and facilitate knowledge transfer, enhancing generalization capabilities. Extensive experiments demonstrate that our method outperforms state-of-the-art LReID methods.

\bibliographystyle{IEEEtran}
\bibliography{DCR.bib}

\vspace{11pt}
\vspace{-33pt}
\begin{IEEEbiography}[{\includegraphics[width=1in,height=1.25in,clip,keepaspectratio]{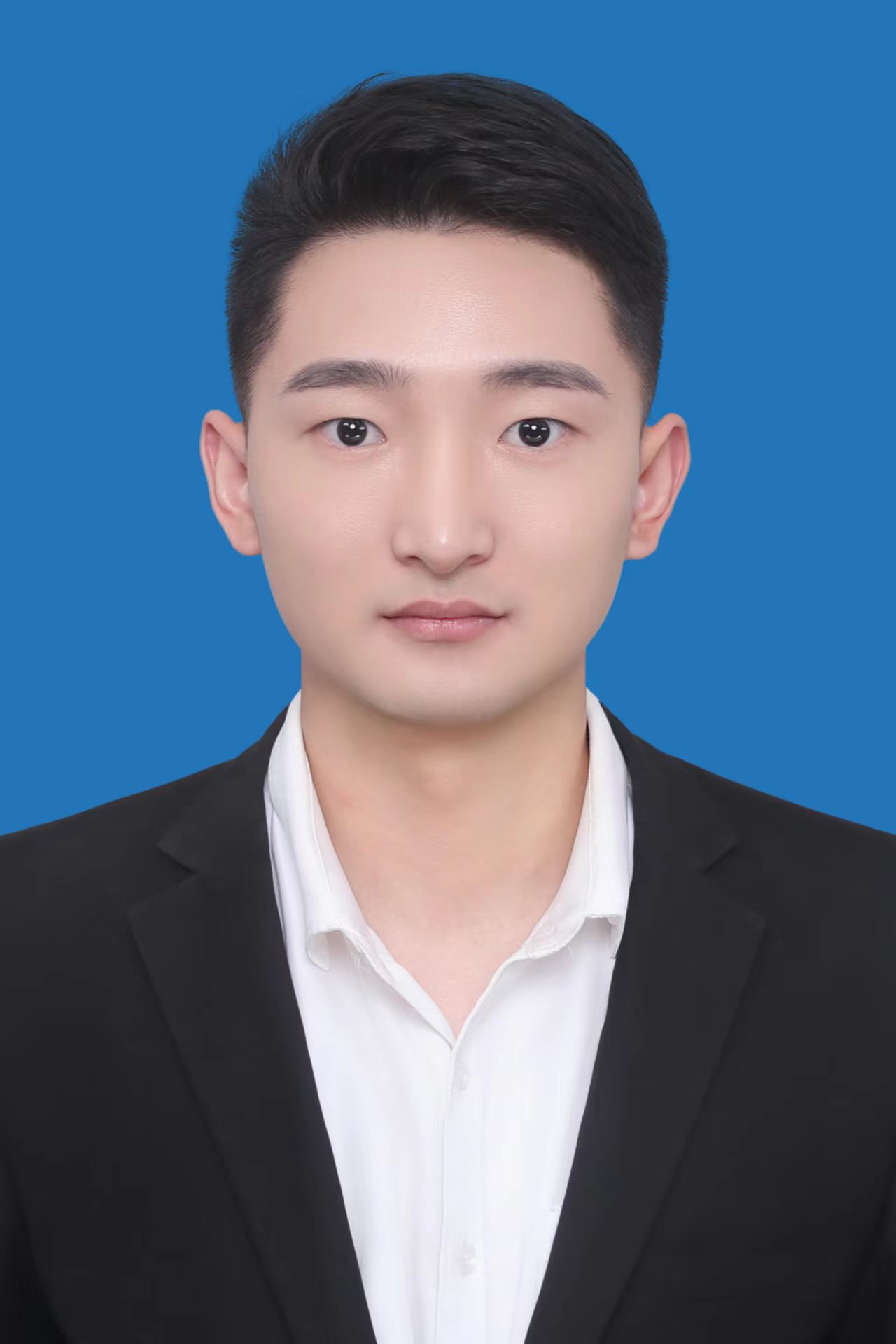}}]{Shiben Liu} received his B.E. and M.S. degrees in Electronic Information Engineering, and Communication and Information Systems from Liaoning University of Engineering and Technology, Fuxin, China, in 2019 and 2022, respectively. He is currently pursuing a Ph.D. degree in mode recognition and intelligent systems from the Chinese Academy of Sciences University, Beijing, China. His current research focuses on deep learning, lifelong learning, person re-identification, image restoration and analysis. He has served	as the Reviewer for the international journals such as the TCSVT, TMM, RAL, Neurocomputing, et al.	
\end{IEEEbiography}
\vspace{11pt}

\vspace{-33pt}
\begin{IEEEbiography}[{\includegraphics[width=1in,height=1.25in,clip,keepaspectratio]{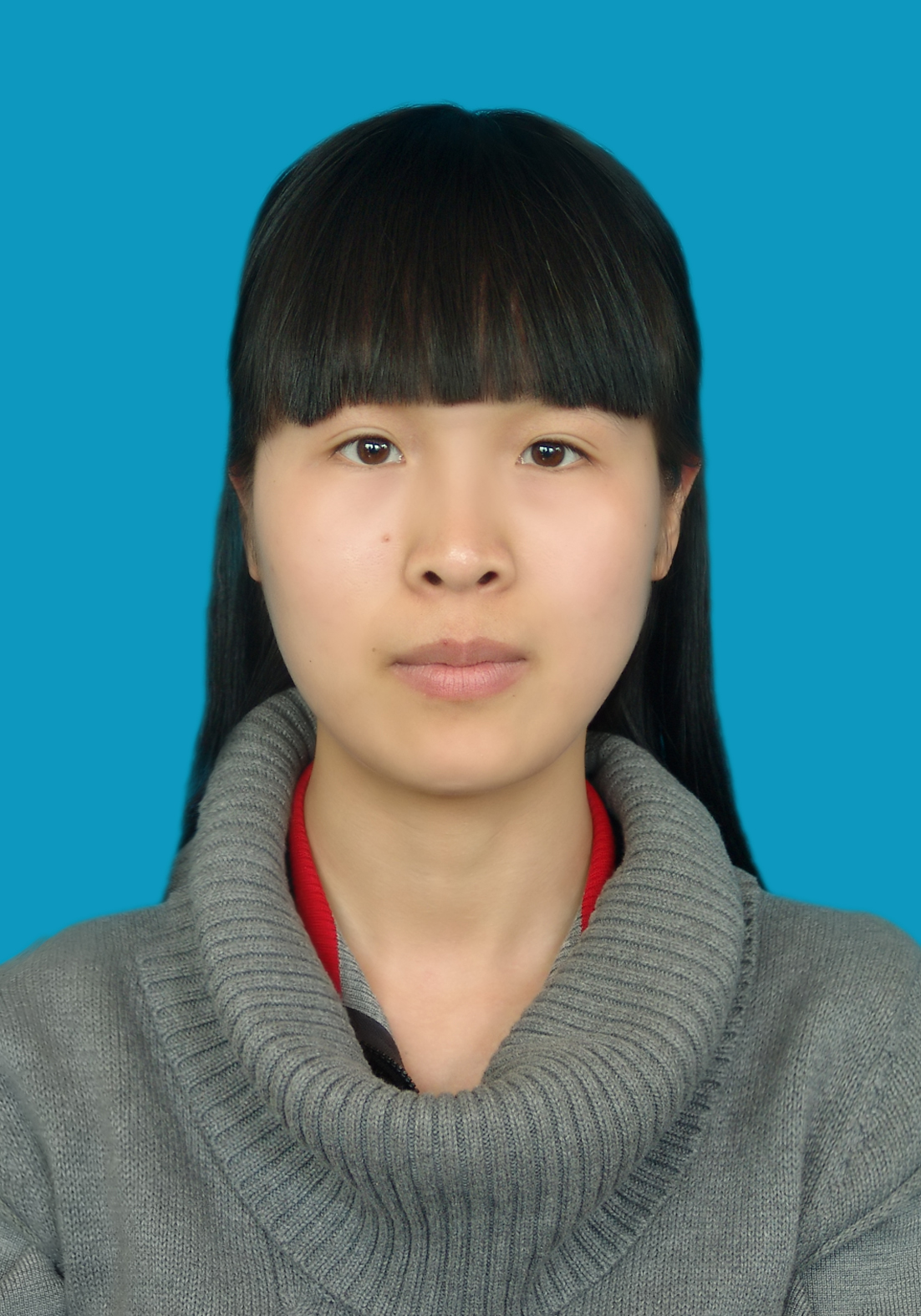}}]{Huijie Fan} (Member, IEEE) received the B.E. degree in automation	from the University of Science and Technology of Science and Technology of China, China, in 2007, and the Ph.D. degree in mode recognition and intelligent systems from the Chinese Academy of Sciences University, Beijing, China, in 2014. She is currently a Research Scientist with the Institute of Shenyang Automation of the Chinese Academy of Sciences. Her research interests include deep learning on image processing and medical image processing and applications.
\end{IEEEbiography}
\vspace{11pt}

\vspace{-33pt}
\begin{IEEEbiography}[{\includegraphics[width=1in,height=1.25in,clip,keepaspectratio]{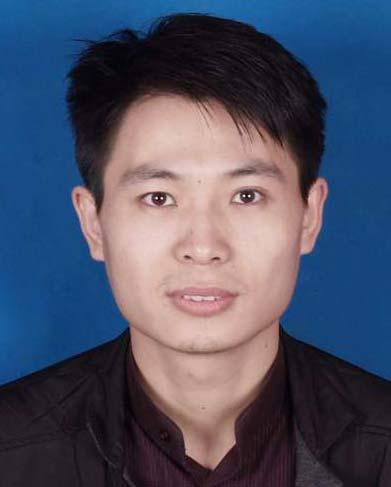}}]{Qiang Wang} received the B.E. and M.S. degrees in school of Computer Science and Technology from Shandong Jianzhu University and Tianjin Normal University, P.R.China,in 2004 and 2008, respectively, the Ph.D degree in State Key Laboratory of Robotics, Shenyang Institute of Automation, University of Chinese Academy of Sciences, Beijing, China in 2020. He is an Associate Professor in the Key Laboratory of Manufacturing Industrial Integrated in Shenyang University. He has some top-tier journal papers accepted at TIP, TMM, TCSVT, IoT-J and Pattern Recognition et al. His current research focuses on deep learning, multi-task learning, image restoration and analysis.
\end{IEEEbiography}
\vspace{11pt}

\vspace{-33pt}
\begin{IEEEbiography}[{\includegraphics[width=1in,height=1.25in,clip,keepaspectratio]{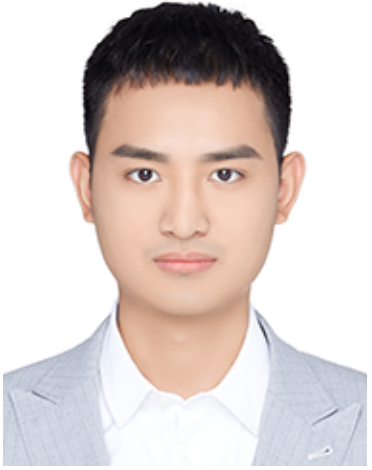}}]{Weihong Ren} received the B.E. degree in Automation and Electronic Engineering from Qingdao University of Science and Technology, Qingdao, China, in 2013. In 2020, he received Ph.D degree from the City University of Hong Kong, Hongkong, China, and the Shenyang Institute of Automation,	Chinese Academy of Sciences, Shenyang, China. He is an Assistant Professor at the School of Mechanical Engineering and Automation, Harbin Institute of Technology, Shenzhen, China. He also has some top-tier journals and conference papers accepted at IEEE TIP, IEEE TMM, CVPR, AAAI et al. His current research interests include object tracking, action recognition and deep learning.
\end{IEEEbiography}
\vspace{11pt}

\vspace{-33pt}
\begin{IEEEbiography}[{\includegraphics[width=1in,height=1.25in,clip,keepaspectratio]{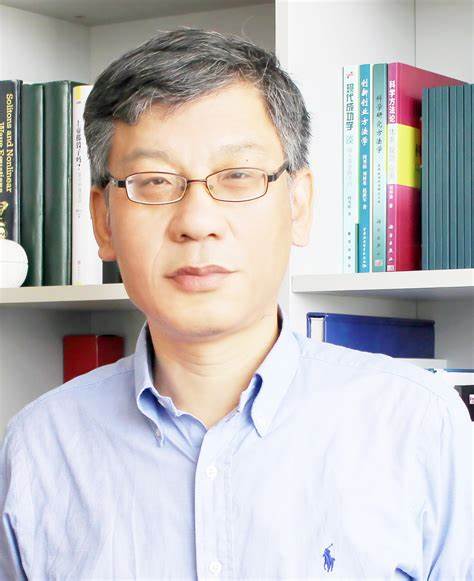}}]{Yandong Tang} (Member, IEEE) received the B.S. and M.S. degree	in the calculation of mathematics from Shandong University, China, in 1984 and 1987. From 1987 to 1996, he worked at the Institute of Computing Technology, Shenyang, Chinese Academy of Sciences. From 1996 to 1998, he was engaged in research and development at Stuttgart University	and Potsdam University in Germany. He received a Ph.D. degree in Engineering Mathematics from the	Research Center (ZETEM) of Bremen University, Germany, in 2002. From 2002 to 2004, he worked at the Institute of Industrial Technology and Work Science (BIBA) at Bremen	University of Germany. He is currently a Research Scientist with the Institute of Shenyang Automation of the Chinese Academy of Sciences. His research interests include image processing, mode recognition and robot vision.
\end{IEEEbiography}

\vspace{11pt}

\vspace{-33pt}
\begin{IEEEbiography}[{\includegraphics[width=1in,height=1.25in,clip,keepaspectratio]{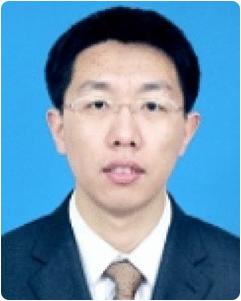}}]{Yang Cong} (Senior Member, IEEE) received the B.Sc. degree from Northeast University in 2004 and the Ph.D. degree from the State Key Laboratory of Robotics, Chinese Academy of Sciences, in 2009. From 2009 to 2011, he was a Research Fellow with the National University of Singapore (NUS) and Nanyang Technological University (NTU). He was a Visiting Scholar with the University of Rochester. He was the professor until 2023 with Shenyang Institute of Automation, Chinese Academy of Sciences. He is currently the full professor with South China University of Technology. He has authored over 80 technical articles. His current research interests include robot, computer vision, machine learning, multimedia, medical imaging and data mining. He has served on the editorial board of the several joural papers. He was a senior member of IEEE since 2015. 
\end{IEEEbiography}

\vfill
\end{document}